\documentclass[10pt,twocolumn,letterpaper]{article}
\usepackage{duckuments}
\usepackage{color}

\usepackage{authblk}

\usepackage{cvpr}              

\usepackage{graphicx}
\usepackage{tabularx}
\usepackage{adjustbox}
\usepackage{listings}
\usepackage{xcolor}
\usepackage{multirow}
\definecolor{cvprblue}{rgb}{0.21,0.49,0.74}
\usepackage[pagebackref,breaklinks,colorlinks,allcolors=cvprblue]{hyperref}

\def\paperID{9010} 
\def\confName{CVPR}
\def\confYear{2025}

\title{Human Motion Instruction Tuning}

\begin{document}



\makeatletter
\def\thanks#1{\protected@xdef\@thanks{\@thanks
        \protect\footnotetext{#1}}}
\makeatother

\author{Lei Li\textsuperscript{\rm 1, 2, *, \dag}, Sen Jia\textsuperscript{\rm 1, *},  Jianhao Wang\textsuperscript{\rm 3}, Zhongyu Jiang\textsuperscript{\rm 1}, Feng Zhou\textsuperscript{\rm 4}, \\ Ju Dai\textsuperscript{\rm 5}, Tianfang Zhang\textsuperscript{\rm 6}, Zongkai Wu\textsuperscript{\rm 1}, Jenq-Neng Hwang\textsuperscript{\rm 1} \thanks{\(^{*}\) These authors contributed equally to this work.} \thanks{\(^{\dag}\) Corresponding Author. (\href{mailto:lilei@di.ku.dk}{\color{black}{\texttt{lilei@di.ku.dk}}}) \\
\textsuperscript{\rm 1}University of Washington \textsuperscript{\rm 2}University of Copenhagen 
\textsuperscript{\rm 3}Xi'an Jiaotong University \textsuperscript{\rm 4}North China University of Technology
\textsuperscript{\rm 5}Peng Cheng Laboratory \textsuperscript{\rm 6}Tsinghua University }
}




\maketitle

\begin{abstract}



This paper presents \textbf{LLaMo} (\textbf{L}arge \textbf{La}nguage and Human \textbf{Mo}tion Assistant), a multimodal framework for human motion instruction tuning. In contrast to conventional instruction-tuning approaches that convert non-linguistic inputs, such as video or motion sequences, into language tokens, LLaMo retains motion in its native form for instruction tuning. This method preserves motion-specific details that are often diminished in tokenization, thereby improving the model’s ability to interpret complex human behaviors. By processing both video and motion data alongside textual inputs, LLaMo enables a flexible, human-centric analysis. Experimental evaluations across high-complexity domains, including human behaviors and professional activities, indicate that LLaMo effectively captures domain-specific knowledge, enhancing comprehension and prediction in motion-intensive scenarios. We hope LLaMo offers a foundation for future multimodal AI systems with broad applications, from sports analytics to behavioral prediction.

\end{abstract}    
\section{Introduction}
\label{sec:intro}

\begin{figure}[ht]
    \centering
    \includegraphics[width=1.0\linewidth]{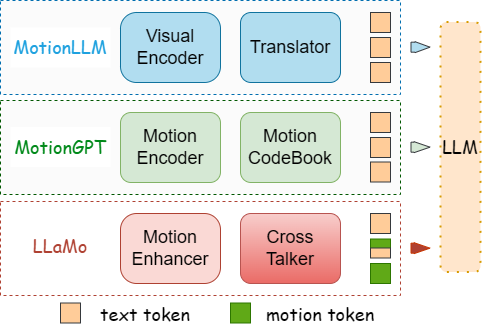}
    \caption{A comparison of MotionLLM~\cite{chen2024motionllm}, MotionGPT~\cite{chen2023motiongpt}, and LLaMo highlights LLaMo's motion-specific capabilities. Equipped with a Motion Enhancer and Cross Talker module to align motion and text, LLaMo supports both video and motion inputs, enabling text-aware, fine-grained motion analysis. }
    \label{fig:Motivation}
\end{figure}

Understanding human motion is a central challenge in multimodal AI, impacting numerous fields such as digital human, human-computer interaction, sports analytics, healthcare, and virtual human modeling~\cite{plappert2016kit, zhang2021we, hong2022versatile,qu2024llms}. Motion data, which captures skeletal movements and body dynamics, provides a structured, appearance-invariant representation of human actions, concentrating on the essential movement patterns while excluding irrelevant visual details~\cite{chen2024dense}. This data has emerged as a privacy-conscious alternative to visual inputs, offering finer control over action recognition and facilitating analysis in applications ranging from digital human avatars to behavioral monitoring in healthcare~\cite{mahmood2019amass, heilbron2015activitynet, lei2025chatmotion}. As digital human representations and motion-centric applications continue to expand, developing models that can leverage the unique characteristics of motion data becomes increasingly crucial for achieving robust, context-aware behavior analysis.

Recently, large language models (LLMs) have shown promise in motion analysis, leveraging their capacity to model temporal dependencies across multimodal data~\cite{chen2024motionllm, endo2023motionqa}. State-of-the-art models, such as MotionGPT~\cite{jiang2023motiongpt} and MotionLLM~\cite{chen2024motionllm}, treat motion data as a multimodal extension, encoding it as discrete tokens or translating it into language-like representations for processing by LLMs. These models have yielded notable results in motion understanding and recognition tasks, illustrating the adaptability of language models to non-linguistic data. However, their reliance on tokenization or textual conversion can limit the richness of the motion representation, abstracting away critical spatial-temporal details essential for high-fidelity human activity recognition.

Despite recent advancements, notable challenges remain in current approaches to motion analysis using LLMs. First, encoding motion data into language tokens, as implemented in models such as MotionGPT, relies on a quantization process that may obscure critical fine-grained motion information, potentially limiting the model’s capacity to accurately capture detailed spatial-temporal dynamics~\cite{shi2023learning, guo2022tm2t}. The common approach of translating motion data into text tokens further contributes to this issue, often resulting in the loss of essential motion-specific nuances that are critical for in-depth behavior interpretation~\cite{li2024mvbench, endo2023motionqa}. This translation process poses considerable challenges in understanding the intricate details of human movement, impeding the model's ability to achieve spatial and physical comprehension of human actions. Moreover, motion data inherently contains 3D structural information and insights into connected human behaviors, which are often difficult to fully capture when relying solely on language tokens, as this representation can overlook the underlying spatial dependencies crucial to modeling human motion accurately.

In addition, existing methods frequently treat video and motion data as isolated modalities, overlooking the potential benefits of a complementary, integrated approach~\cite{an2024multimodality}. Processing these data types independently neglects the rich environmental context and interaction cues provided by video inputs, which are crucial for accurately interpreting complex human behaviors~\cite{mahmood2019amass, hong2022versatile}. This separation limits the capacity for a holistic understanding of human motion, particularly in scenarios where context, spatial relations, and dynamic interactions play essential roles in behavior analysis. Furthermore, video-based approaches, while potentially more informative, are computationally intensive, posing significant constraints for large-scale or real-time applications, which limits their feasibility in practical motion analysis systems.

In response to these challenges, we propose LLaMo, a framework designed to integrate motion data as a distinct modality within LLMs without translating it into textual intermediaries. \textbf{LLaMo incorporates a \textbf{Motion Estimator} and an \textbf{Enhancer} to facilitate the direct input of motion or/and video data, thereby offering a unified, human-centric analysis pipeline.} The framework also introduces a \textbf{Cross Talker} module, which dynamically aligns the motion and text features, allowing for text-aware, fine-grained motion representations. This approach retains the original spatial-temporal characteristics of the motion data, providing a more precise understanding of human actions and enabling high-resolution behavior analysis in motion-intensive contexts~\cite{bodenheimer1997motion, peng2024text}. 

Through extensive evaluations on benchmarks such as MoVid-Bench~\cite{chen2024motionllm} and BABEL-QA~\cite{endo2023motion}, we demonstrate that LLaMo achieves state-of-the-art performance in motion-centric human activity recognition~\cite{endo2023motionqa, li2024mvbench}. These results underscore LLaMo's potential to advance multimodal AI, providing a robust foundation for human behavior understanding. Our contributions can be summarized as follows:
\begin{itemize}
    \item We introduce LLaMo, which treats motion data as an independent modality within LLMs, preserving essential motion-specific details for robust analysis.
    \item We propose \textbf{Cross Talker}, a text-guided mechanism that dynamically focuses on and aggregates key motion frames features, optimizing computational efficiency and enabling the model to pick relevant motion features.
    \item LLaMo’s architecture supports the direct input of both raw motion and video data, providing a generalized framework for human-centric analysis across diverse applications, such as sports analytics, healthcare, and behavioral monitoring.
\end{itemize}

\section{Related Work}

\subsection{Human-Centric Multimodal Representation}

Multimodal representation learning is integral to human-centric analysis, especially for applications requiring spatial-temporal reasoning to decode complex behaviors~\cite{lin2023videollm, ning2023videobench, li2023videochat}. Recent advances, such as Video-LLaVA, have made significant strides by embedding visual data from images and videos into a shared linguistic feature space, enabling sophisticated visual reasoning for behavioral analysis tasks~\cite{lin2023videollm, li2023segment}. Despite these achievements, many current models are optimized primarily for static images or discrete video frames, limiting their effectiveness in sequential, dynamic scenarios where understanding progression and continuity in motion is crucial~\cite{ning2023videobench, heilbron2015activitynet, maaz2023video}. Privacy concerns further challenge the adoption of video-based methods in sensitive domains; thus, researchers are exploring motion data as a privacy-conscious alternative, enabling analysis focused on human actions without revealing identifiable visual information~\cite{song2023adaptive, yang2023recognizing, li2023mask}. By combining visual and motion data, emerging multimodal frameworks hold promise for more comprehensive, privacy-aware human behavior analysis, leveraging the strengths of both modalities for adaptability and depth across applications.

\subsection{Human Motion Understanding}

Human motion understanding has traditionally relied on skeletal data represented as sequences of joint keypoints to capture movement while preserving user privacy~\cite{shi2023learning, plappert2018bidirectional,li2024cpseg,yang2023understanding}. Early models like 2s-AGCN~\cite{shi2019two} and more recent transformer-based methods, such as MotionCLIP~\cite{chen2023motiongpt}, have achieved success in tasks like activity recognition, captioning, and behavior analysis by mapping motion data to language tokens, effectively capturing structural aspects of movement. However, these methods often lack environmental context, which is crucial for nuanced interpretations in real-world applications, as similar movements can convey different meanings across varied scenarios~\cite{song2023finegrained, maaz2023video, an2023temporal}. To address this limitation, recent frameworks integrate motion and visual data, allowing models to generalize across diverse, dynamic environments~\cite{liu2024pose, he2023activitynet, li2023hierarchical}. This fusion of skeletal data with video-provided context has shown promise in applications like sports and professional activity analysis, where contextual cues are essential. Additionally, skeletal-based data remains valuable in privacy-sensitive applications, such as healthcare and security, where it allows for robust analysis without compromising personal identity.

\begin{figure*}[t]
    \centering
    \includegraphics[width=\textwidth]{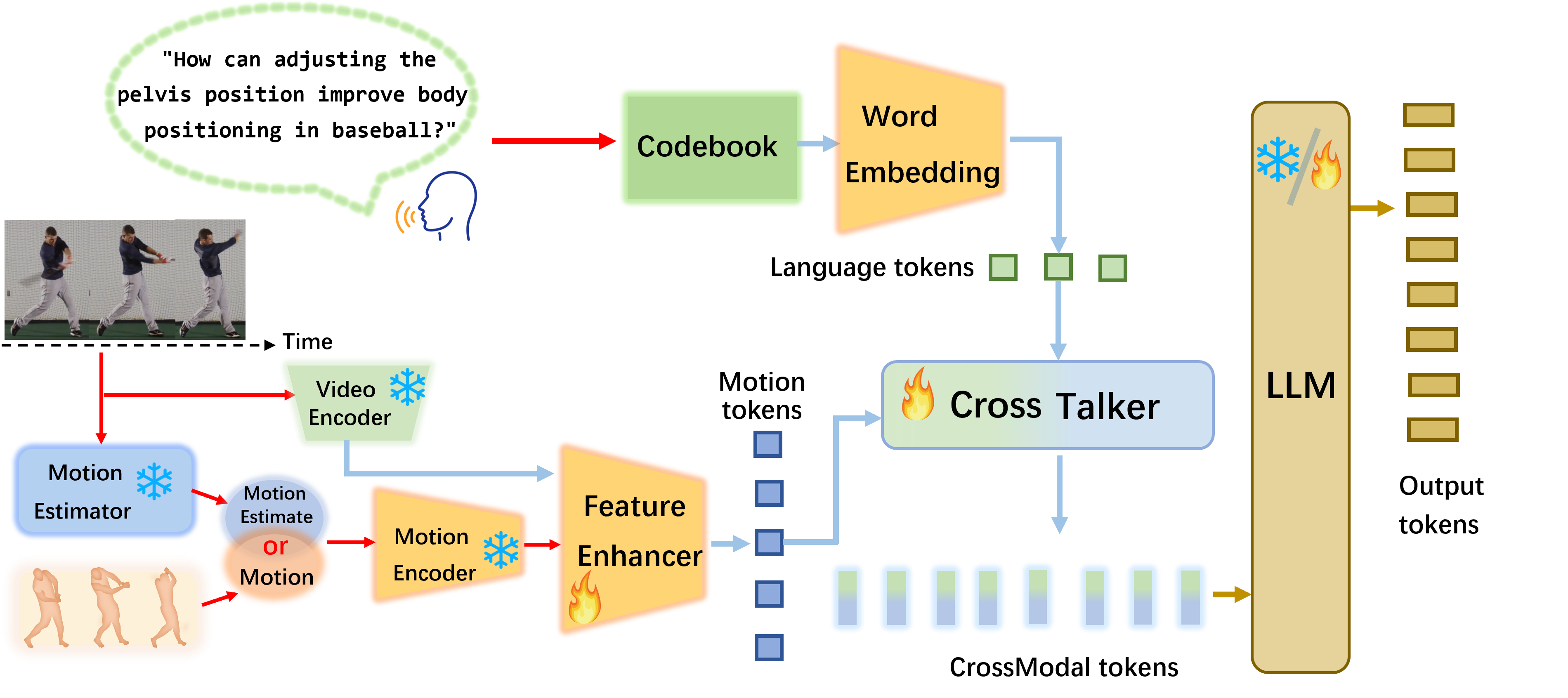}
    \caption{Overview of the LLaMo framework. It includes three main modules: (1) Multimodal Feature Extraction for encoding video and motion data; (2) Cross Talker for aligning and fusing motion and text features; and (3) Behavior Generation Module to produce text descriptions of human behavior based on integrated features.}
    \label{fig:Main_pipeline}
\end{figure*}
\section{Methods}

The LLaMo framework is composed of three primary modules to effectively process and integrate video, motion data and text. An overview of this pipeline is provided in Figure~\ref{fig:Main_pipeline}. The architecture comprises the following components:

\begin{enumerate}
    \item \textbf{Multimodal Feature Extraction Module}: This module independently encodes the video or motion inputs, which makes LLaMo is a general human motion assistant with video input or motion input. Then the Motion features are enhanced by distilling valuable representations in video.
    
    \item \textbf{Cross Talker}: Here, the enhanced motion features are aggregated in a text-guided manner and aligned with text within a shared semantic space, allowing the natural feature fusion between motion and text, which provides fine-grained inputs for LLM.
    
    
    \item \textbf{Behavior Generation Module}: Using the aggregated and text-aligned motion features, along with the text representations, this module generates a contextually aware textual description of the observed human behavior.
\end{enumerate}

\subsection{Enhanced motion Feature Extraction Module}

The enhanced motion feature extraction module consists of four primary components: the motion estimator, followed by the motion and video encoders, and the feature enhancement module. The motion estimator addresses scenarios where motion data may be unavailable by estimating motion information directly from video frames. The motion and video encoders then independently process the motion and video data, encoding them into their respective feature spaces. This approach allows the system to leverage complementary information from both modalities, creating a richer representation of human behavior.

Given a sequence of video frames \( V = \{v_1, v_2, \dots, v_T\} \), the video encoder \( f_v(\cdot) \) extracts a set of visual features:
\[
F_V = f_v(V) = \{f_v(v_1), f_v(v_2), \dots, f_v(v_T)\}.
\]
Simultaneously, the motion encoder \( f_m(\cdot) \) processes the motion data \( M = \{m_1, m_2, \dots, m_T\} \) to generate motion-specific features:
\[
F_M = f_m(M) = \{f_m(m_1), f_m(m_2), \dots, f_m(m_T)\}.
\]
Here, \( F_V \in \mathbb{R}^{T \times H} \) and \( F_M \in \mathbb{R}^{T \times H} \) represent the extracted feature sets for the video and motion data, respectively, where \( T \) is the sequence length and \( H \) denotes the feature dimension.

The feature enhancement module then aligns and fuses these modality-specific features. By using the motion features \( F_M \) as queries, this module selectively extracts semantically relevant information from the video features \( F_V \). This alignment and fusion process allows the motion features to be enriched with contextual information from the video, producing refined, contextually enriched representations that enhance the model’s understanding of nuanced human behaviors. For the detailed design refer to the Appendix.



First, the video features and motion features will conduct self-attention operations respectively to get their augmented features $F_V^{\prime}$ and $F_M^{\prime}$. This allows the video and motion to initially capture the dependencies on their respective time steps. Then, in order to enrich the motion data features with semantic information from the video, we subtly design a cross-attention block where augmented motion features as the query to extract valid semantic information in the video features. Finally, the selected video semantic representation and motion features are connected by residuals followed by a feed-forward neural network to obtain the enhanced motion features $\tilde{F_M}$

\subsection{Cross Talker Module}

\begin{figure}[ht]
    \centering
    \includegraphics[width=1\columnwidth]{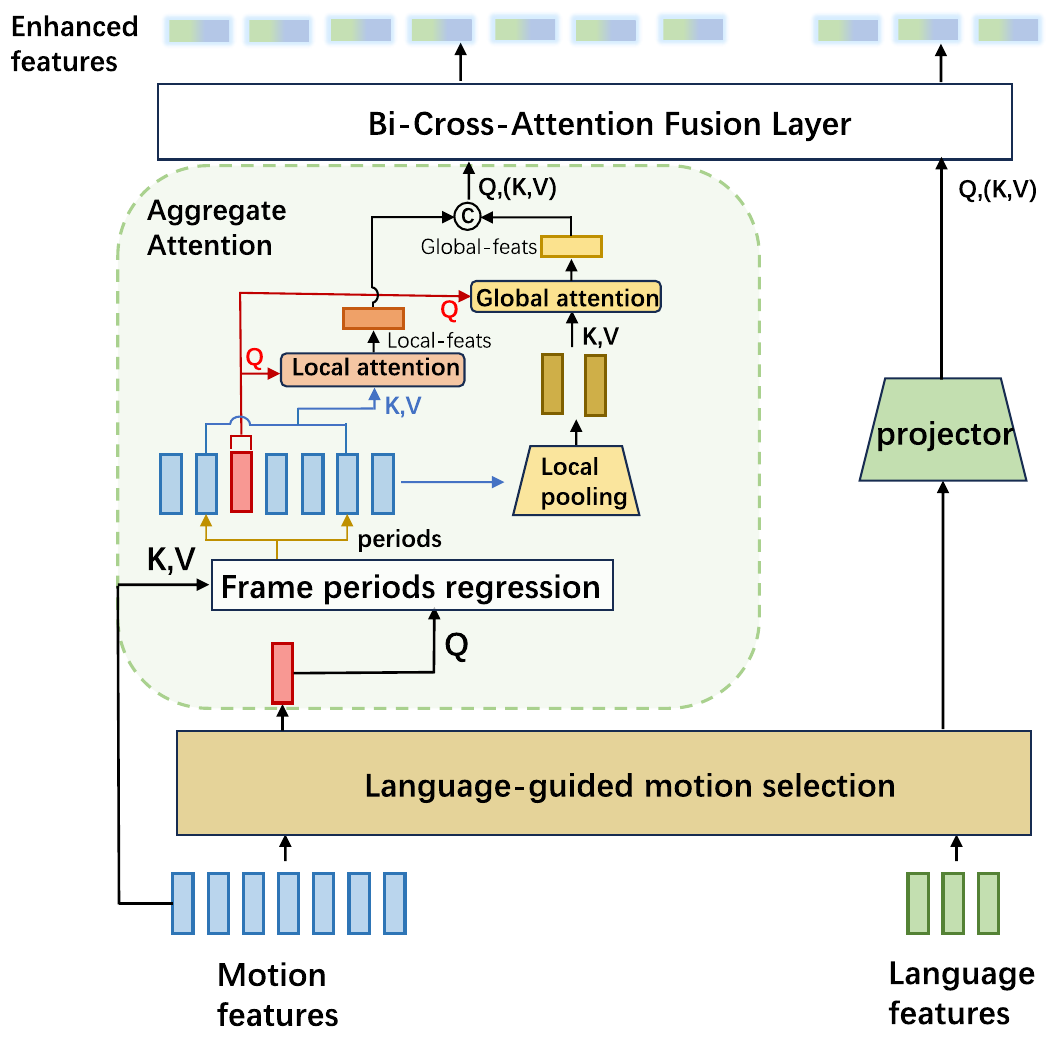} 
    \caption{Overview of the Cross Talker Module, which selects key frames based on text guidance and fuses them with text features for enhanced analysis.}
    \label{fig:Cross_talker}
\end{figure}

Given the enhanced motion features $\tilde{F}_M$ from the feature enhancement module, directly feeding the entire sequence of motion frames into the Large Language Model (LLM) incurs significant computational costs and may lead to suboptimal performance. This limitation arises from the self-attention mechanism in transformers, which has a computational complexity of $\mathcal{O}(L^2)$, where $L$ denotes the sequence length. When incorporating both motion and text, the sequence length becomes $L = L_T + T$, where $L_T$ is the number of text tokens and $T$ is the number of motion frames. As $T$ increases, the quadratic growth in computational cost becomes prohibitive, leading to complexity proportional to $(L_T + T)^2$.

Furthermore, computing attention over long sequences can dilute attention weights, making it challenging for the model to focus on essential parts of the sequence and effectively capture long-range dependencies. Furthermore, the motion features are not well aligned with text either. To address these issues, we propose the Cross Talker Module, which selectively distills key motion frames—referred to as \emph{viewpoint frames}—based on their relevance to the text context, as well as aligning the motion features with the text. An overview of the Cross Talker Module is shown in Figure~\ref{fig:Cross_talker}.

\paragraph{Language-Guided Frame Selection} Given the enhanced motion features $\tilde{F}_M \in \mathbb{R}^{T \times H}$ and text embeddings $F_T \in \mathbb{R}^{L_T \times H}$, we aim to identify $K$ viewpoint frames that are most relevant to the textual context. This selection is achieved through a cross-attention operation, where text features act as queries and motion features as keys and values:
\begin{equation}
A = \text{Softmax}\left(\frac{F_T W_Q (\tilde{F}_M W_K)^\top}{\sqrt{d}}\right),
\end{equation}
where $W_Q, W_K \in \mathbb{R}^{H \times d}$ are learnable projection matrices, and $d$ is the dimensionality of the queries and keys. The resulting attention matrix $A \in \mathbb{R}^{L_T \times T}$ contains attention weights, with each element $A_{i, j}$ indicating the relevance between the $i$-th text token and the $j$-th motion frame.

To determine the importance of each motion frame, we aggregate attention weights across all text tokens using max pooling:
\begin{equation}
s_j = \max_{i=1, \dots, L_T} A_{i, j}, \quad j = 1, \dots, T.
\end{equation}
The resulting scores $s_j$ represent the significance of each frame with respect to the text context. We then select the top $K$ frames with the highest relevance scores as viewpoint frames, reducing the effective sequence length from $T$ to $K$. This reduction in sequence length lowers the self-attention complexity from $\mathcal{O}((L_T + T)^2)$ to $\mathcal{O}((L_T + K)^2)$, which is computationally efficient when $K \ll T$.

\paragraph{Adaptive Contextual Feature Aggregation} For each selected viewpoint frame, we enrich its representation by aggregating both local and global contexts. An adaptive receptive field size $r_k$ is predicted for each viewpoint frame $k$ using a receptive field regression module. The module performs a cross-attention operation where each viewpoint frame serves as a query, and the unselected motion features serve as keys and values, followed by a sigmoid activation to estimate $r_k$. Based on $r_k$, we define a local window $W_k$ around frame $k$: 
\begin{equation}
W_k = \{j \mid |j - k| \leq r_k \times T\}.
\end{equation}

Within this window, we apply local attention to capture fine-grained details:
\begin{equation}
F_{\text{local}}(k) = \tilde{F}_M(k) + \text{Attention}(\tilde{F}_M(k), \tilde{F}_M(W_k), \tilde{F}_M(W_k)),
\end{equation}
where $\tilde{F}_M(W_k)$ denotes the motion features within the local window around frame $k$.

To incorporate global context, we partition the full motion sequence into $N$ segments and compute segment-level features via average pooling:
\begin{equation}
F_M^{\text{seg}} = \{\bar{F}_M(n)\}_{n=1}^N,
\end{equation}
where the segment size is controlled by the hyperparameter $S_n$. We then apply global attention:
\begin{equation}
F_{\text{global}}(k) = F_{\text{local}}(k) + \text{Attention}(F_{\text{local}}(k), F_M^{\text{seg}}, F_M^{\text{seg}}).
\end{equation}
The final representation for each viewpoint frame is obtained by concatenating local and global features:
\begin{equation}
F_M(k) = [F_{\text{local}}(k); F_{\text{global}}(k)] \in \mathbb{R}^{2H},
\end{equation}
allowing the model to capture both fine-grained motion details and broader contextual information.

\paragraph{Bidirectional Cross-Modal Fusion} To integrate the enhanced motion features with the text embeddings, we employ a bidirectional cross-attention mechanism. First, we update the motion features by performing a cross-attention operation, where the motion features act as queries and the text features serve as keys and values, and vice versa for the text features. The updated features are then passed through feed-forward networks, and we concatenate the motion and text representations to form the input for the next stage:
\begin{equation}
F_{\text{fusion}} = [F_T; \{F_M(k)\}_{k \in K}].
\end{equation}

This bidirectional fusion enhances the interaction between motion and text, enabling a coherent, contextually grounded understanding of human behavior. By selectively focusing on key motion frames and integrating enriched motion and text features, the Cross Talker Module reduces computational costs and improves the model’s ability to capture complex human actions.


\subsection{Behavior Generation Module}

The Behavior Generation Module leverages the fused features \( F_{\text{fusion}} \) to generate a textual description \( Y \) that encapsulates the observed human behavior. This module utilizes a language model \( h(\cdot) \) to transform \( F_{\text{fusion}} \) into a semantically coherent and contextually relevant output sequence:
\begin{equation}
Y = h(F_{\text{fusion}}) = \{ y_1, y_2, \dots, y_L \},
\end{equation}
where \( y_t \) denotes the token generated at the \( t \)-th time step, and \( L \) is the length of the generated description. The probability of generating each token \( y_t \) is given by:
\begin{equation}
p(y_t \mid y_{1:t-1}, F_{\text{fusion}}) = \text{Softmax}(W_o h_t),
\end{equation}
where \( h_t \) represents the hidden state of the language model at time \( t \), and \( W_o \) is an output projection matrix. The hidden state \( h_t \) is derived from the model’s internal mechanisms, which integrate attention over the input features and previously generated tokens. This autoregressive approach enables LLaMo to produce detailed and contextually aligned descriptions that incorporate nuanced information from both motion and text modalities.

By iteratively generating tokens based on the fused features and preceding context, the Behavior Generation Module constructs a meaningful and comprehensive narrative of human activity, grounded in the multimodal input data.

\subsection{Training Objective}

To train LLaMo, we finetune a pretrained language model using supervised learning, guiding it to generate accurate and contextually relevant descriptions based on the input motion, video, and text features. Both the motion and video encoders remain frozen during training, focusing the optimization on the language model’s alignment with multimodal inputs. The training objective minimizes the discrepancy between the generated descriptions and the ground-truth annotations in the dataset.

Given a training set of \( N \) samples, each with video frames \( V^{(i)} \), motion data \( M^{(i)} \), and corresponding ground-truth textual descriptions \( \hat{Y}^{(i)} = \{ \hat{y}^{(i)}_1, \hat{y}^{(i)}_2, \dots, \hat{y}^{(i)}_L \} \), the overall training objective is to minimize the negative log-likelihood of the ground-truth tokens across all samples:
\begin{equation}
L = -\frac{1}{N} \sum_{i=1}^{N} \sum_{t=1}^{L^{(i)}} \log p(\hat{y}^{(i)}_t \mid \hat{y}^{(i)}_{1:t-1}, F_{\text{fusion}}^{(i)}),
\end{equation}
where \( F_{\text{fusion}}^{(i)} \) represents the fused input features for the \( i \)-th sample. The term \( p(\hat{y}^{(i)}_t \mid \hat{y}^{(i)}_{1:t-1}, F_{\text{fusion}}^{(i)}) \) denotes the probability of the ground-truth token \( \hat{y}^{(i)}_t \) given the previous tokens \( \hat{y}^{(i)}_{1:t-1} \) and the fused features.

This objective ensures that the generated descriptions are syntactically and semantically aligned with the target outputs, capturing the detailed aspects of human behavior present in the video and motion inputs.

\section{Experiments}

\begin{figure*}[t]
   \centering
   \includegraphics[width=\linewidth]{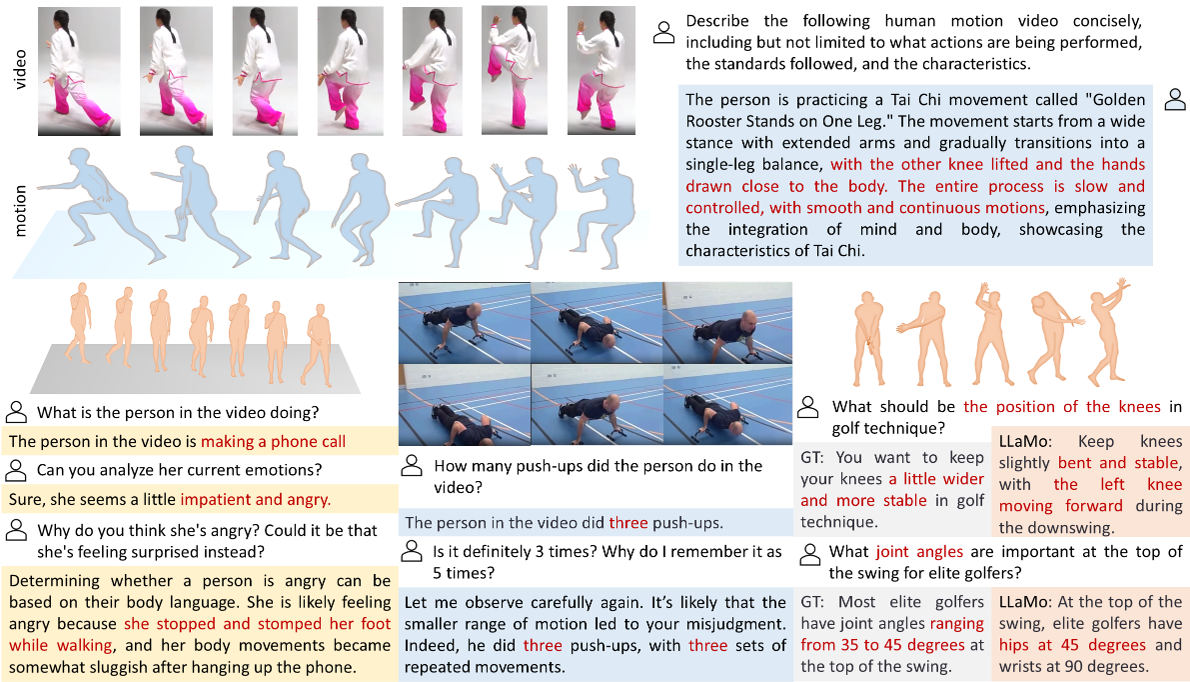} 
   \caption{Example outputs from LLaMo across human activities and professional sports, showcasing its reasoning capabilities and domain-specific knowledge in motion-intensive scenarios.}
   \label{fig:outputs_examples}
\end{figure*}

To evaluate LLaMo, we performed experiments across datasets representing complex human behaviors and tasks, including professional sports and cyclic action repetition counts.

\subsection{Implementation Details}

\paragraph{Training Datasets:} LLaMo was trained using a combination of bimodal video-motion datasets and unimodal video or motion datasets to enable a robust understanding of both modalities. For video-motion data, we utilized MoVid from MotionLLM~\cite{chen2024motionllm} and our custom Swing dataset containing 20,000 annotated videos of baseball and golf swings with corresponding motion capture data and expert Q\&A instructions. Motion-only datasets included HumanML3D~\cite{guo2022generating} and KIT-ML~\cite{plappert2016kit}, while Mo-RepCount, a filtered subset of RepCount~\cite{hu2022transrac}, provided high-quality repetition count videos. This multimodal data configuration ensures that LLaMo effectively learns from both single and combined modalities.

\paragraph{Evaluation Datasets:} LLaMo was evaluated on MoVid-Bench~\cite{chen2024motionllm} to assess both video and motion understanding capabilities, with additional tests on BABEL-QA~\cite{endo2023motion} to evaluate motion-based question answering. For specific tasks, we used Mo-RepCount to gauge motion details and Swing to evaluate professional sports guidance. 

\paragraph{Evaluation Metrics:} For MoVid-Bench, we follow previous LLM evaluation metrics~\cite{lin2023video}~\cite{li2023videochat}~\cite{jin2024chat} on accuracy and scores, which contains body-part awareness, sequentially, direction
analysis, reasoning ability, and hallucination, respectively. BABEL-QA used the metric followed by \cite{endo2023motion}, Mo-RepCount applied metrics such as OBO, MAE, OBZ, and RMSE, and Swing performance was assessed on reasonableness, coherence, pertinence, and adaptability using GPT-4 evaluations, scoring from 0 to 5.

\subsection{Results}

\begin{table*}[t]
\centering

\renewcommand{\arraystretch}{0.8} 
\caption{Expected Comparison on the MoVid-Bench. The top part of the table presents motion-related results, and the bottom part presents video-related results. Higher accuracy and score values indicate better performance.}
\label{table:expected_results}
\small

\resizebox{\textwidth}{!}{ 

\begin{tabular}{l|cc|cc|cc|cc|cc|cc}
\toprule
\textbf{MoVid-Bench-Motion} & \multicolumn{2}{c|}{\textbf{Body.}} & \multicolumn{2}{c|}{\textbf{Seq.}} & \multicolumn{2}{c|}{\textbf{Dir.}} & \multicolumn{2}{c|}{\textbf{Rea.}} & \multicolumn{2}{c|}{\textbf{Hall.}} & \multicolumn{2}{c}{\textbf{All}} \\
& \textbf{Acc.} & \textbf{Score} & \textbf{Acc.} & \textbf{Score} & \textbf{Acc.} & \textbf{Score} & \textbf{Acc.} & \textbf{Score} & \textbf{Acc.} & \textbf{Score} & \textbf{Acc.} & \textbf{Score} \\ 
\midrule
GT & 100.00 & 5.00 & 100.00 & 5.00 & 100.00 & 5.00 & 100.00 & 5.00 & 100.00 & 5.00 & 100.00 & 5.00 \\
GPT-3.5~\cite{openai2023gpt35} & 24.51 & 2.04 & 30.41 & 2.25 & 27.14 & 2.19 & 39.19 & 2.64 & 58.33 & 3.22 & 31.33 & 2.31 \\
MotionGPT~\cite{chen2023motiongpt} & 31.22 & 3.98 & 42.69 & \textbf{3.16} & 44.29 & 3.50 & 35.81 & 3.06 & 16.66 & 2.25 & 36.86 & 3.11 \\
MotionLLM~\cite{chen2024motionllm} & 50.49 & 3.55 & 36.84 & 3.14 & 58.57 & 3.76 & 52.70 & 3.58 & 55.56 & 3.39 & 49.50 & 3.49 \\ 
\textbf{LLaMo} & \textbf{59.30} & \textbf{4.01} & \textbf{44.01} & 3.12 & \textbf{60.91} & \textbf{3.99} & \textbf{58.21} & \textbf{3.64} & \textbf{61.17} & \textbf{3.53} & \textbf{55.32} & \textbf{3.67} \\ 
\midrule

\textbf{MoVid-Bench-Video} & \multicolumn{2}{c|}{\textbf{Body.}} & \multicolumn{2}{c|}{\textbf{Seq.}} & \multicolumn{2}{c|}{\textbf{Dir.}} & \multicolumn{2}{c|}{\textbf{Rea.}} & \multicolumn{2}{c|}{\textbf{Hull.}} & \multicolumn{2}{c}{\textbf{All}} \\
& \textbf{Acc.} & \textbf{Score} & \textbf{Acc.} & \textbf{Score} & \textbf{Acc.} & \textbf{Score} & \textbf{Acc.} & \textbf{Score} & \textbf{Acc.} & \textbf{Score} & \textbf{Acc.} & \textbf{Score} \\ 
\midrule
GT & 100.00 & 5.00 & 100.00 & 5.00 & 100.00 & 5.00 & 100.00 & 5.00 & 100.00 & 5.00 & 100.00 & 5.00 \\
GPT-3.5~\cite{openai2023gpt35} & 2.40 & 1.23 & 1.39 & 1.00 & 4.65 & 1.09 & 5.41 & 1.65 & 0.00 & 0.94 & 3.03 & 1.26 \\
Video-LLAVA~\cite{lin2023video} & 33.53 & 2.76 & 25.46 & 2.72 & 41.86 & 2.84 & 52.97 & 3.28 & 58.83 & 1.89 & 42.53 & 2.70 \\
MotionLLM~\cite{chen2024motionllm} & \textbf{34.13} & \textbf{2.93} & 32.87 & 2.92 & 44.18 & 3.14 & 63.20 & 3.55 & 70.59 & \textbf{2.30} & 49.00 & 2.97 \\ 
\textbf{LLaMo} & 33.83 & 2.85 & \textbf{36.01} & \textbf{3.11} & \textbf{45.50} & \textbf{3.32} & \textbf{67.59} & \textbf{3.73} & \textbf{72.81} & 2.25 & \textbf{52.33} & \textbf{3.10} \\ 
\bottomrule
\end{tabular}
}
\end{table*}

\paragraph{Evaluation on Motion Understanding in MoVid-Bench.} We evaluated LLaMo's motion understanding capabilities on the MoVid-Bench, focusing on five critical aspects. Both accuracy and score metrics were used to comprehensively assess LLaMo's performance. Table~\ref{table:expected_results} provides a comparison with several baselines, including GPT-3.5~\cite{openai2023gpt35}, MotionGPT~\cite{chen2023motiongpt}, and MotionLLM~\cite{chen2024motionllm}. The GPT-3.5 baseline, limited to textual data processing, struggles significantly with motion interpretation, yielding low performance. MotionGPT, while capable of incorporating motion data, is primarily code motion into a codebook, which limits its reasoning ability on coiled motion data and leaves it vulnerable to hallucinations. Although MotionLLM provides improvements in understanding motion sequences, it still conducts translation on motion data, resulting in low accuracy on seq. and Hall.

LLaMo, in contrast, achieves superior results across nearly all metrics on MoVid-Bench-Motion, attaining the highest accuracy in most categories. This performance is primarily attributed to its preservation of critical motion-specific nuances, as LLaMo processes motion data as a distinct modality without requiring any conversion. LLaMo effectively captures the intricate details of 3D sequences associated with human behaviors. Its Feature Enhancer, combined with the Cross Talker module, enables simultaneous input of video and motion data, allowing LLaMo to selectively focus on motion frames most relevant to the text context. This design enhances the model's ability to capture complex patterns of human motion and improves its reasoning on motion-intensive tasks. Overall, LLaMo's results on MoVid-Bench demonstrate its effectiveness in advancing human behavior analysis, highlighting its potential for broader applications in general motion understanding.

\begin{table*}[t]
\centering
\renewcommand{\arraystretch}{0.8} 
\setlength{\tabcolsep}{7pt} 
\caption{Comparison on BABEL-QA dataset. Higher scores indicate better performance.}
\label{table:babel_qa_comparison}
\small
\begin{tabular}{l|c|cccc|ccc}
\toprule
\textbf{Model} & \textbf{Pred. type} & \textbf{Overall} $\uparrow$ & \textbf{Action} $\uparrow$ & \textbf{Direction} $\uparrow$ & \textbf{Body Part} $\uparrow$ & \textbf{Before} $\uparrow$ & \textbf{After} $\uparrow$ & \textbf{Other} $\uparrow$ \\ 
\midrule
MotionCLIP-M~\cite{openai2023gpt35} & cls. & 0.430 & 0.485 & 0.361 & \textbf{0.272} & 0.372 & 0.321 & 0.404\\ 
MotionCLIP-R~\cite{openai2023gpt35} & cls. & 0.420 & 0.489 & 0.310 & 0.250 & 0.398 & 0.314 & 0.387\\ 
MotionLLM~\cite{openai2023gpt35} & gen. & 0.436 & 0.517 & 0.354 & 0.154 & 0.427 & 0.368 & \textbf{0.529} \\ 
\midrule 
\textbf{LLaMo} & gen. & \textbf{0.458} & \textbf{0.525} & \textbf{0.398} & 0.224 & \textbf{0.443} & \textbf{0.392} & 0.518 \\ 
\bottomrule
\end{tabular}
\label{tab:BABEL-QA}
\end{table*}

\begin{table*}[t]
\centering
\renewcommand{\arraystretch}{0.8} 
\setlength{\tabcolsep}{10pt} 
\caption{Performance on the profession-swing dataset across four indicators. Higher accuracy and score values indicate better performance.}
\label{table:golf_swing_results}
\small
\begin{tabular}{l|cc|cc|cc|cc|cc}
\toprule
\textbf{Model} & \multicolumn{2}{c|}{\textbf{Reasonableness}} & \multicolumn{2}{c|}{\textbf{Coherence}} & \multicolumn{2}{c|}{\textbf{Pertinence}} & \multicolumn{2}{c|}{\textbf{Adaptability}} & \multicolumn{2}{c}{\textbf{All}} \\
& \textbf{Acc} & \textbf{Score} & \textbf{Acc} & \textbf{Score} & \textbf{Acc} & \textbf{Score} & \textbf{Acc} & \textbf{Score} & \textbf{Acc} & \textbf{Score} \\
\midrule
GT & 100 & 5 & 100 & 5 & 100 & 5 & 100 & 5 & 100 & 5 \\
GPT-3.5~\cite{openai2023gpt35} & 4.51 & 0.83 & 4.20 & 0.74 & 3.65 & 0.66 & 1.71 & 0.22 & 3.42 & 0.61 \\
MotionGPT~\cite{chen2023motiongpt} & 10.53 & 1.25 & 15.42 & 1.66 & 12.64 & 1.1 & 14.79 & 1.35 & 14.35 & 1.4 \\
MotionLLM~\cite{chen2024motionllm} & 12.33 & 1.51 & 19.20 & 1.87 & 17.98 & 1.79 & 10.20 & 1.22 & 16.53 & 1.57 \\
\midrule 
\textbf{LLaMo} & \textbf{21.10} & \textbf{2.11} & \textbf{27.10} & \textbf{2.71} & \textbf{31.81} & \textbf{3.12} & \textbf{20.22} & \textbf{1.98} & \textbf{24.80} & \textbf{2.48} \\
\bottomrule
\end{tabular}
\end{table*}

\paragraph{Evaluation on Video Understanding in MoVid-Bench.} We evaluated LLaMo on the video component of the MoVid-Bench dataset to assess its capability for comprehensive video understanding. As presented in Table~\ref{table:expected_results}, LLaMo achieves strong performance across most metrics, demonstrating LLaMo's extraordinary comprehension ability on videos even without motion input.

Compared to other models like GPT-3.5~\cite{openai2023gpt35}, Video-LLAVA~\cite{lin2023video}, and MotionLLM~\cite{chen2024motionllm}, while LLaMo slightly trails behind MotionLLM in body-part awareness, LLaMo shows significant improvements in reasoning ability and overall scores. This indicates that our model excels in understanding complex video content and making accurate inferences about human behaviors.

The rewarding results can be attributed to our model's powerful ability to estimate motion data accurately in video and align it seamlessly with video inputs. This approach retains the motion capture of motion data and the auxiliary information provided by video, enriching the understanding of dynamic scenes.

\paragraph{Evaluation on BABEL-QA.} We evaluated LLaMo on the BABEL-QA dataset to assess its capability to handle complex motion-based queries. As shown in Table~\ref{tab:BABEL-QA}, LLaMo achieves the highest overall score of 0.458, outperforming all baselines. It excels particularly in the \textit{Action} category with a score of 0.525, demonstrating superior action recognition capabilities. Additionally, it achieves top scores in temporal reasoning tasks (\textit{Before} and \textit{After}), indicating a strong understanding of temporal relationships in motion sequences. This capability benefits from the direct integration of motion data as an independent modality and the effective text-aware motion frames extract mechanism in LLaMo, preserving crucial motion-specific nuances, as well as leading to more accurate and contextually rich interpretations.

\paragraph{Professional Sports Analysis.} We further evaluate LLaMo on the challenging swing-dataset, termed as profession-swing, details refer to Appendix, where the ground truth (GT) consists of answers provided by professional coaches, serving as expert benchmarks. The evaluation focuses on four key indicators: \textbf{Reasonableness}: Logical soundness and plausibility of the responses. \textbf{Coherence}: Consistency and logical flow within the responses. \textbf{Pertinence}: Relevance of the answers to the specific questions. \textbf{Adaptability}: Ability to adjust responses based on the stage on the athlete. 

As presented in Table~\ref{table:golf_swing_results}, LLaMo significantly outperforms both GPT-3.5~\cite{openai2023gpt35} and MotionLLM~\cite{chen2024motionllm} across all metrics. While MotionLLM shows improvements over GPT-3.5, with higher accuracy and scores, it still lags behind LLaMo. Specifically, LLaMo achieves an overall accuracy of 24.80

The superior performance of LLaMo can be attributed to the design of the Cross Talker module, which enables language-guided frame selection to capture complex motion relationships inherent in professional swings. This allows LLaMo to effectively model intricate motion patterns and generate responses that are more reasonable, coherent, pertinent, and adaptable, closely aligning with the expert-level insights provided by professional coaches.

In contrast, while MotionLLM demonstrates certain strengths over MotionGPT—such as better pertinence and coherence. It lacks the advanced motion understanding capabilities of LLaMo, lagging behind LLaMo distinctly.

\begin{table}[h!]
\caption{Motion and video details capture evaluation on Mo-RepCount}
\small 
\centering
\begin{tabular}{p{2cm}|p{0.8cm} p{0.8cm} p{0.8cm} p{0.8cm}}
\toprule
\textbf{Model}   & \textbf{OBO} & \textbf{MAE} & \textbf{OBZ} & \textbf{RMSE} \\ \midrule
EScounts~\cite{sinha2024every}         & \textbf{0.397}        & \textbf{0.291}        & 0.198        & \textbf{5.58}          \\
PoseRAC~\cite{yao2023poserac}         & 0.382        & 0.312        & 0.204        & 5.95          \\
TransRAC~\cite{hu2022transrac}         & 0.276        & 0.444        & 0.105        & 8.56          \\
RepNet~\cite{wandt2019repnet}           & 0.009        & /            & /            & /             \\ \midrule
\textbf{LLaMo} & 0.389 & 0.324 & \textbf{0.222} & 6.15 \\ \bottomrule
\end{tabular}
\label{table:mo-repcount-evaluation}
\end{table}

\paragraph{Evaluation on Mo-RepCount.} We evaluated LLaMo on the Mo-RepCount dataset to assess its ability to capture fine-grained motion details and complex temporal features required for accurate repetition counting. For a fair comparison, we trained all state-of-the-art models, including RepNet~\cite{wandt2019repnet}, TransRAC~\cite{hu2022transrac}, PoseRAC~\cite{yao2023poserac} and EScounts~\cite{sinha2024every}, on the Mo-RepCount dataset without using any additional data.  As presented in Table~\ref{table:mo-repcount-evaluation}, LLaMo outperforms the RepNet and TransRAC across almost all metrics. Since both of them are models built for loop counting tasks, the effect of our model on this task is convincing, which shows that LLaMo is also competent in capturing the details of the motion and videos. 

While LLaMo did not outperform the specialized counting models EScounts and poseRAC on many metrics, it excelled on the OBZ metric, demonstrating an unmatched ability to capture fine-grained motion details and spatio-temporal relationships. These results confirm LLaMo’s strong potential in motion and video analysis, underscoring its suitability for detail-oriented tasks and applications.

\section{Conclusion}

We introduce LLaMo, a novel framework that integrates motion data directly into large language models, thereby enhancing multimodal understanding. By unifying text, video, and motion data in an architecture fine-tuned for human motion instruction, LLaMo achieves notable progress in behavior comprehension, motion-focused captioning, and interactive Q\&A, capturing the fine-grained motion details vital for precise, context-aware insights. Experimental results, especially in professional sports analysis, demonstrate LLaMo’s capacity to interpret complex human motion with accuracy. Future work will extend LLaMo to a broad range of domains where detailed motion analysis and multimodal understanding are essential. The main challenges lie in refining cross-modal integration techniques and improving efficiency for real-time applications, thus broadening LLaMo’s potential in human-centered AI. We hope our work will spur further exploration into more advanced human-centric multimodal models.

\section*{Acknowledgement}
This work was supported in part by the Pioneer Centre for AI, DNRF grant number P1.



{
    \small
    \bibliographystyle{ieeenat_fullname}
    \bibliography{main}
}


\end{document}




\input{sec/X_suppl.tex}
{
    \small
    \bibliographystyle{ieeenat_fullname}
    \bibliography{main}
}